\documentclass{bmvc2k}
\usepackage{amssymb}
\usepackage{booktabs}
\usepackage{multirow}

\usepackage{hyperref}
\usepackage{gensymb,graphics,subfigure,inputenc}
\usepackage[linesnumbered,algo2e,boxed]{algorithm2e}
\usepackage[figuresright]{rotating}
\usepackage{textcomp,subfigure,multirow,upgreek}
\usepackage{mdwlist}
\usepackage{amsmath}

\usepackage{ragged2e}
\usepackage{fontawesome}
\graphicspath{{images/}}

\usepackage{geometry}
\usepackage{siunitx}
\usepackage{booktabs, makecell, multirow, tabularx}
\usepackage{lipsum}

\newcommand\blfootnote[1]{%
  \begingroup
  \renewcommand\thefootnote{}\footnote{#1}%
  \addtocounter{footnote}{-1}%
  \endgroup
}
\usepackage{caption}
\usepackage{enumitem}
\setitemize{noitemsep,topsep=0pt,parsep=0pt,partopsep=0pt}

\title{AniFormer: Data-driven 3D Animation with Transformer}

\addauthor{Haoyu Chen}{chen.haoyu@oulu.fi}{1}
\addauthor{Hao Tang}{hao.tang@vision.ee.ethz.ch}{2}
\addauthor{Nicu Sebe}{nicu.sebe@unitn.it}{3}
\addauthor{Guoying Zhao*}{guoying.zhao@oulu.fi}{1}
\addinstitution{
CMVS\\
University of Oulu
}
\addinstitution{
 Computer Vision Lab\\
 ETH Zurich
}

\addinstitution{
 DISI\\
 University of Trento
}

\runninghead{H. CHEN, H. Tang, N. Sebe, G. Zhao}{AniFormer}

\begin{document}

\maketitle

\begin{abstract}
\noindent
We present a novel task, i.e., animating a target 3D object through the motion of a raw driving sequence. In previous works, extra auxiliary correlations between source and target meshes or intermedia factors are inevitable to capture the motions in the driving sequences. Instead, we introduce AniFormer, a novel Transformer-based architecture, that generates animated 3D sequences by directly taking the raw driving sequences and arbitrary same-type target meshes as inputs. Specifically, we customize the Transformer architecture for 3D animation that generates mesh sequences by integrating styles from target meshes and motions from the driving meshes. Besides, instead of the conventional single regression head in the vanilla Transformer, AniFormer generates multiple frames as outputs to preserve the sequential consistency of the generated meshes. To achieve this, we carefully design a pair of regression constraints, i.e., motion and appearance constraints, that can provide strong regularization on the generated mesh sequences. Our AniFormer achieves high-fidelity, realistic, temporally coherent animated results and outperforms compared start-of-the-art methods on benchmarks of diverse categories. Code is available: \url{https://github.com/mikecheninoulu/AniFormer}.
\end{abstract}

\section{Introduction}
\label{sec:intro}
\blfootnote{* Indicates the corresponding author}
Directly animating 3D objects via dynamics from raw driving sequences is a desirable task. Apart from purely scientific interests, learning to automatically synthesize high-fidelity 3D animation has a wide range of applications. For example, in the film industry, an automatic 3D animating model finds use indirectly transferring motions to various 3D characters without any need of manual designing the kinetic procedures. Using a learned 3D animating model in virtual reality (VR) scenarios, one can generate realistic animated 3D avatars without explicitly specifying the geometry properties and dynamics, which would be cumbersome but inevitable when using a standard graphics rendering engine.

There are various existing works related to 3D animation problem, including 3D mesh deformation \cite{skeletondeformation,NPT,Unsupervised,LIMP,vaelimp,dt,3dcode,clothdeformation}, 3D motion estimation \cite{vibe,vibenew,meshgrapher}, and 3D motion synthesis \cite{vaesynthesis,motionsynthesis}. In this paper, we study a new form, i.e., data-driven 3D animation. At the core, we aim to learn a mapping function that can animate a target 3D object driven by the motion from a raw mesh sequence. To the best of our knowledge, a general-purpose solution to raw-data driven 3D animation has not yet been explored by prior work, although its similar tasks, the 3D deformation and style transfer problems have been intensively researched that could also achieve 3D animation by processing single mesh frame by frame \cite{clothdeformation,NPT,Unsupervised,dt}. Our method is inspired by previous 2D video synthesis works \cite{firstorder,vid2vid}.

We cast the raw data-driven 3D animation as a continuous deformation problem, where the goal is to train a model such that the animated object has the desired dynamics from driving sequences while preserving the original appearance information. To this end, we utilize the strength of sequential attention mechanism from Transformers to learn given paired input and output 3D sequences. Specifically, we design a new Transformer structure that can jointly learn the style information from target meshes and dynamics from the driving sequences. To handle sequential mesh processing, we replace the classical regression head with various regression constraints for appearance and motion consistency over the whole generated sequences. Our method can learn to generate high-fidelity, realistic, and temporally coherent 3D human motions. Conditioning on the same driving sequences, our model can produce identical 3D motions for various seen and unseen target 3D human meshes. Moreover, we extend our method to different objects such as animals.

\begin{figure}[!t] \small
    \centering
    \includegraphics[width=0.8\textwidth]{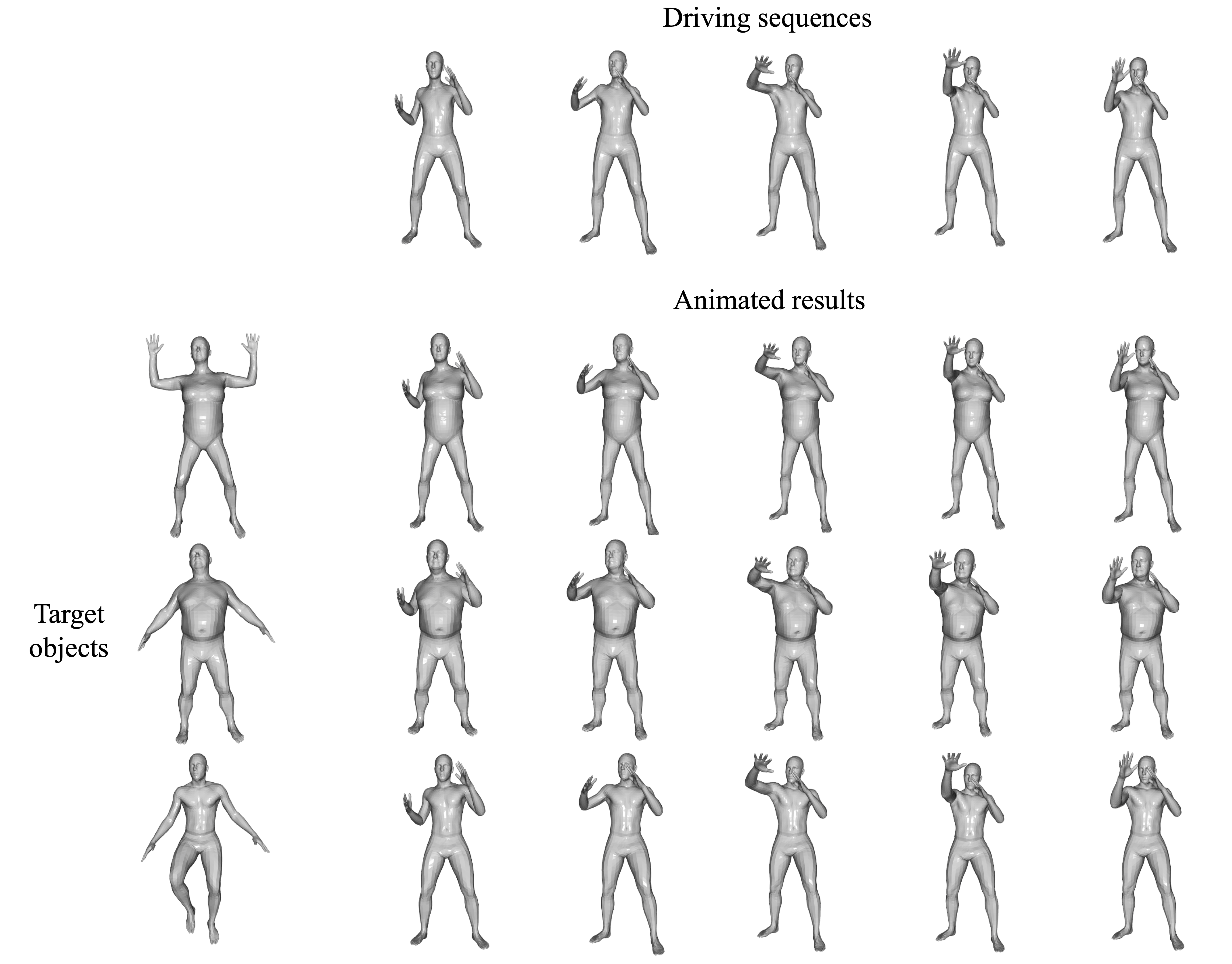}
    \caption{Example produced animations from AniFormer: a driving sequence (top), different target objects (left), and animated resulting sequences (bottom). The samples are from DFAUST dataset \cite{DFAUST}. The motion is directly learnt from raw meshes of the driving sequences without extra manual interventions.}
    \vspace{-0.4cm}
    \label{fig:transformer comparison}
\end{figure}

In summary, our main contributions are:
\begin{itemize}[leftmargin=*]
\item To our knowledge, our work is the first that can achieve end-to-end 3D animation that learns the 3D motion directly from raw data sequences without any extra intervention.
\item We propose the AniFormer, a Transformer-based architecture for 3D animation, which integrates both styles and motions from source meshes. It also discards the convention regression head with a single-vector representation and directly generates multiple frames as outputs to better learn the sequential consistency.
\item Two regression constraints are carefully designed to synthesize realistic animated results, i.e., the natural appearance consistency in the driving sequences, and motion consistency between the driving sequence and synthesized sequence.
\item Experimental results on four different datasets demonstrate that the proposed method achieves the state-of-the-art performance. Moreover, we further represent the generalization capability by extending our framework to diverse categories, such as various SMPL human meshes \cite{SMPL}, non-SMPL human meshes, and animals, and showing it is a strong baseline for data-driven 3D animation.
\end{itemize}

\section{Related Work}

\textbf{3D Mesh Deformation} aims to generate a new 3D shape with a given pair of source and target shapes, which can be seen as a frame-level task of 3D animation. Towards 3D mesh deformation, previous works demand re-enforcing the correspondence between source and target meshes. For example, some disentanglement-based methods like \cite{LIMP,Unsupervised} use the shape correspondences between different pose meshes from the same body to decompose shape and pose factors. SNARF \cite{skeletondeformation} uses differentiable forward skinning for animating non-rigid neural implicit shapes, which can be extended to unseen and complex shapes, but bone transformations between the target meshes and the desired pose are critical in their settings. SCANimate \cite{clothdeformation} learns a pose-aware parametric clothed human model by introducing an intermediate canonicalization procedure to convert the posed space into canonical space, however, its animation is achieved by manipulating SMPL pose parameters but not driven by raw data sources. As mentioned, all these existing deformation methods need extra intervened information which is non-trivial, and time-consuming to obtain. In contrast, our framework does not require any additional auxiliary information and can animate unseen target meshes with completely different source mesh sequences. Moreover, unlike existing works \cite{IEPGAN}, we are the first to achieve the 3D animation with long raw sequential driven data. 

\noindent\textbf{3D Motion with Sequential Dependencies.} Generating 3D mesh sequences via motion information has been intensively researched from different aspects. For 3D mesh pose and motion estimation, VIBE \cite{vibe} and TCMR \cite{vibenew} extract static features from the sequential frames into a temporal feature by using a bidirectional gated recurrent unit (GRU) \cite{gru}. To synthesize 3D body meshes with high fidelity, they use an SMPL \cite{SMPL} parameter regressor to provide strong constraints on the synthesized meshes. For 3D motion synthesis, ACTOR \cite{vaesynthesis} and Action2Motion \cite{action2motion} learn generators based on \cite{vaesource} that achieve action-conditioned generation of realistic and diverse 3D human mesh sequences. Again, an SMPL joint regressor is introduced to ACTOR to ensure the realistic generated motions and Action2Motion post-renders the skeleton joints to SMPL models to obtain better visualized results. Compared to the above works, we have two core differences: 1) both the appearance of target meshes and desired motions can be handled in the generated 3D sequences, and 2) our framework directly processes the raw mesh sequences without any manual intervention (e.g., SMPL parameters or skeleton joints) in an end-to-end manner.

\noindent\textbf{Transformer for Temporal Modeling.} Transformer architecture was
first proposed by \cite{transformer} and raised emerging research interests in vision tasks with promising performances \cite{transsurvey}. To retain positional information of sequential dependencies, a trainable linear projection layer to embed each input patch to a high dimensional feature is always applied in, such as pose estimation \cite{poseformer}, motion synthesis \cite{vaesynthesis}, facial analysis \cite{yu2021transrppg}, and video analysis \cite{vit}. However, we argue that, the 3D animation task is different from the above tasks, i.e., its appearance consistency and motion continuity is crucial information that can be utilized as good regression constraints. Meanwhile, the regression head in the vanilla Transformer is in a single-vector representation which is hard to preserve the sequential consistency. Thus, we propose a novel architecture, the AniFormer, based solely on the existing generating flows, dispensing with the conventional regression head entirely for the 3D animation task.

\section{Data-Driven 3D Animation}
Similar to the work of \cite{vid2vid,firstorder} for the 2D task, we are interested in animating a target 3D object based on the motion of a driving sequence. This section first presents a general introduction of the whole AniFormer framework. Then, we will demonstrate how to use AniFormer to achieve 3D animation learning with proposed constraints.

\subsection{AniFormer: Transformer-Based Network for 3D Animation}
\label{sec:GCtransformer}
\begin{figure*}[]
    \centering
    \includegraphics[width=\textwidth]{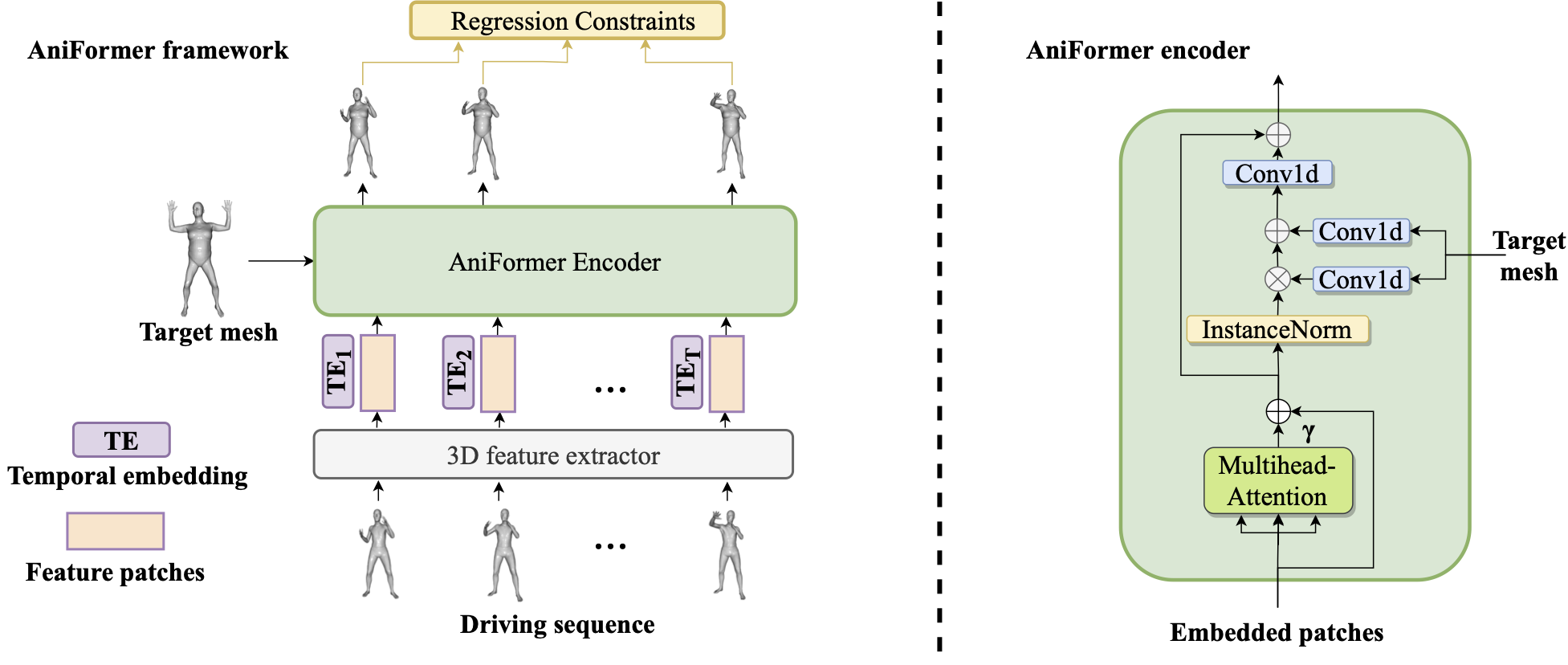}
    \caption{Model overview. We process the driving sequential meshes into patches, add temporal embeddings, and feed the resulting sequence of vectors to our AniFormer encoder. AniFormer encoder can adaptively generate deformed target meshes with given pose meshes via attention mechanism. The illustration was inspired by ViT \cite{vit}.}
    \label{fig:overview}
\end{figure*}

In this proposed animation task, the latent representations of 3D models should be large enough to preserve the detailed surface geometry of the target mesh. However, this will make the number of learnable parameters of conventional recurrent temporal models like RNNs grow quadratically with the memory size, making them extremely large and almost non-trainable. Thus we propose to implement Transformer framework with simple temporal embedding for saving memory. An overview of the AniFormer is depicted in Fig.~\ref{fig:overview}. Our AniFormer consists of three key components, i.e., a 3D mesh feature extractor, an AniFormer encoder, and regression constraints following the generated data flow. 

\noindent\textbf{3D Feature Extractor.} Given a 3D mesh driving sequence of $T$ frames $M_1, \cdots, M_T$, where $M \in \mathbb{R}^{V \times 3}$ and $V$ stands for vertex number such as 6,890, a feature extractor based on the classical PointNet~\cite{pointnet} extracts a static 3D feature per frame, resulting a sequence of latent embedding vector $Z_1, \cdots, Z_T$, where $Z \in \mathbb{R}^{C \times V}$ and $C = 1,024$ in practice. The network weights of the feature extractor are shared for all frames. From the extracted static features of all input 3D meshes, we retain the current frame’s temporal information of the sequence via the classical positional embeddings used in Transformer networks. The procedure can be formulated as:
\begin{equation}
\label{Definemath}
\begin{aligned}
Z = \{Z_1; Z_2; \cdots; Z_T\} + E_{t}.
\end{aligned}
\end{equation}
After summing with the temporal embedding $E_t \in \mathbb{R}^{T \times C \times V }$, the resulting feature vector $Z$ is sent to the AniFormer encoder. In our case, original batch normalization layers are replaced with Instance Normalization~\cite{instance} layers to preserve the instance style~\cite{styletransfer,tang2020local,spatiallynormalization,tang2020dual}.

\noindent\textbf{AniFormer Encoder.} The core problem of casting Transformer networks for animating target 3D meshes is how to jointly learn the correlations between the given target mesh and driving mesh sequences, since Transformer networks are all designed for self-attention mechanism with single-source inputs. To this end, inspired by style transfer works \cite{NPT, styletransfer,spatiallynormalization}, we design the novel AniFormer encoder by integrating the style information from the target 3D mesh into the self-attention mechanism. Specifically, as shown in Fig.~\ref{fig:overview} right part, we feed the embedded latent embedding vector of the pose meshes as $Z$ into different 1D convolution layers to generate the normalized representations $\mathbf{q,k,v}$ for the standard multihead self-attention \cite{transformer}. Then, the attention weights $A_{i,j}$ based on the pairwise geometry between $\mathbf{q}$ and $\mathbf{k}$ is given with the following formula:
\begin{equation}
\label{geometric correlation}
\mathbf{A}_{i,j} = \frac{exp(\mathbf{q}_{i}\mathbf{k}_{j})}{\sum_{i=1}^{n}exp(\mathbf{q}_{i}\mathbf{k}_{j})}.
\end{equation}
After this, a matrix multiplication between $v$ and the transpose of $\mathbf{A}$ is conducted. Finally, we weigh the result with a scale parameter $\gamma$ and conduct an element-wise sum operation with the original latent embedding $Z$ to obtain the refined latent embedding $Z'$,
\begin{equation}
\label{updateembedding}
Z' =\gamma \sum_{i=1}^{n}\left ( \mathbf{A}_{i,j}\mathbf{v}_{i} \right ) + Z,
\end{equation}
where $\gamma$ is a trainable parameter and initialized as 0. After obtaining the latent embedding vector $Z'$, instead of stacking MLP blocks to the $Z'$ as a vanilla Transformer does, the target mesh $M_{target}$ as an external input is fed into two different 1D convolution layers to produce the modulation parameters which are multiplied and added to the $Z'$, as shown in Fig. \ref{fig:overview} right part. In such an explicit-imposing way, our AniFormer encoder can capture sequential motion information and producing a more robust animation of the target mesh. More detailed network architecture and implementation are illustrated in Supplementary Materials. 

Note that, different than previous Transformer-based networks~\cite{poseformer,transformer,vit,yang2021transformer}, our AniFormer could not only compute the attention weights on the embedded patches but also could fully preserve the local geometric details of the target mesh. Besides, by implementing the self-attention mechanism, the AniFormer can take pure latent feature vectors as inputs to drive the target meshes, while in the work of NPT \cite{NPT}, extra concatenation operators are needed for better deformation. Most importantly, our AniFormer is designed for sequential 3D mesh data processing which has never been attempted in those works.

\noindent\textbf{From Regression Head to Regression Constraints.} In almost all the existing Transformer-based networks, a regression head with an MLP block and one linear layer is used to take the average in the sequential dimension and produce the final output, such as image classification, pose estimation. However, we argue that this simple regression head might not apply to the task of 3D animation (also see more experimental results in Section \ref{sec:experiments}), because a regression head with single output cannot provide constraints of appearance and motion consistency over continuous frames, which is crucial for high fidelity 3D animation. To this end, we discard the conventional regression head and enable AniFormer to generate multiple frames by enforcing carefully designed regression constraints over those frames as introduced in the following section.

\subsection{Imposing Regression Constraints}
Our AniFormer is based on Transformer architectures and allows end-to-end training with raw mesh sequences as inputs. However, directly inheriting the regression head in the vanilla Transformer will lead to a degenerated performance since it aggregates all the sequential information on one target without consistency constraints along the temporal dimension. To this end, we discard the regression head and design two regression constraints to ensure that the generated sequences have both motion consistency to the driving sequences and appearance consistency to the target mesh.

\noindent\textbf{Constraint on Motion Consistency.} During training, we can constrain the generated meshes to have the identical kinetic motion using a simple $L1$ loss on the driving mesh sequence. However, due to the geometric gap between the target mesh and driving meshes, their motion trajectories are not identical. Thus, directly implementing the $L1$ loss would cause a geometric distortion problem and not lead to our expected results. To this end, we design a normalized regulation to preserve the motion consistency:
\begin{equation}
\label{motionloss}
\begin{aligned}
\mathcal{L}_{m} = \frac{1}{T-2}\sum_{t=3}^{T}\left(\frac{\| M_{t-1} - M_{t-2}\|^{2}_{2}}{\| G_{t-1} - G_{t-2}\|^{2}_{2}}-\frac{\| M_{t} - M_{t-1}\|^{2}_{2}}{\| G_{t} - G_{t-1}\|^{2}_{2}}\right),
\end{aligned}
\end{equation}
where $M$ stands for generated sequential meshes and $G$ stands for the standard motion mesh from the driving sequences. As mentioned, the absolute motion trajectory of target meshes and driving meshes are not equal due to their different geometric proportions. But using our designed Eq. \eqref{motionloss}, we can proportionally compare the kinetic trajectories and achieve an efficient constraint on the motion of the generated 3D animations.

\noindent\textbf{Constraint on Appearance Consistency.} We introduce an appearance regularisation term as a constraint, which penalises the local distortion between the generated meshes and given target mesh. This regularization is inspired by the edge loss from \cite{3dcode} that can enforce the generated mesh surface to be tight, resulting in a smooth surface with local triangles identical to target meshes. Specifically, we generalize the edge loss into sequential settings, let $\mathcal{N}(\mathbf{p})$ denotes the neighbor vertices of vertex $\mathbf{p}$ and $V$ is the total vertex number of the mesh, the appearance regularization can be defined as below:
\begin{equation}
\label{apploss}
\begin{aligned}
\mathcal{L}_{a} = \frac{1}{T}\frac{1}{V}\sum_{t=1}^{T}\sum_{\mathbf{p}}\sum_{\mathbf{u}\in \mathcal{N} (\mathbf{p})} \left(\left \|\mathbf{p}_M^{t} -\mathbf{u}_M^{t}\|^{2}_{2}-\| \mathbf{p}_N - \mathbf{u}_N\right \|^{2}_{2}\right),
\end{aligned}
\end{equation}
where $M$ and $N$ stand for the generated sequential meshes and the target mesh, respectively.

\noindent\textbf{Full Objective Function.} At last, we define the full objective function as below:
\begin{equation}
\label{loss}
\mathcal{L}_{full} =\lambda_{r}\mathcal{L}_{r} + \lambda_{m}\mathcal{L}_{m} + \lambda_{a}\mathcal{L}_{a},
\end{equation}
where $\mathcal{L}_{r}$, $\mathcal{L}_{m}$, and $\mathcal{L}_{a}$ are the three used losses, including reconstruction loss, motion loss, and appearance loss. $\lambda$ is the corresponding weight of each loss. In Eq.~\eqref{loss}, reconstruction loss $\mathcal{L}_{r}$ is a simple point-wise $L2$ distance commonly used in previous works \cite{NPT,Unsupervised,LIMP}.

\section{Experiments}
\label{sec:experiments}

\begin{figure*}[!t] \small
    \centering
    \includegraphics[width=0.75\textwidth]{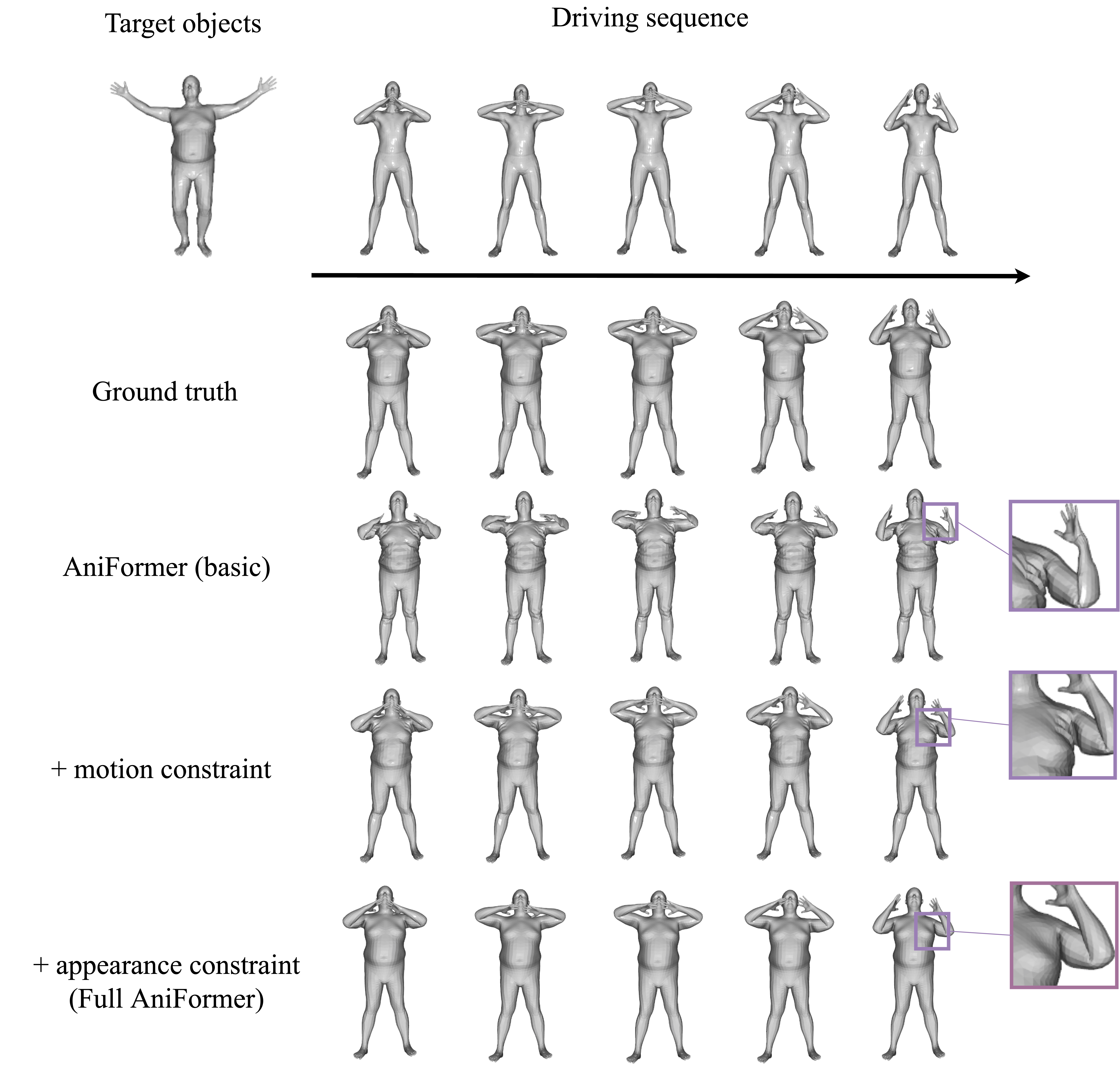}
    \caption{Visualized ablation study by progressively enabling each component of our model with last line as the full AniFormer. As shown, the basic AniFormer can already enable the animation while the constraints makes it more plausible with less distortions.}
    \vspace{-0.4cm}
    \label{fig:transformer ablation}
\end{figure*}

\begin{figure}[]\small
    \centering
    \includegraphics[width=0.8\textwidth]{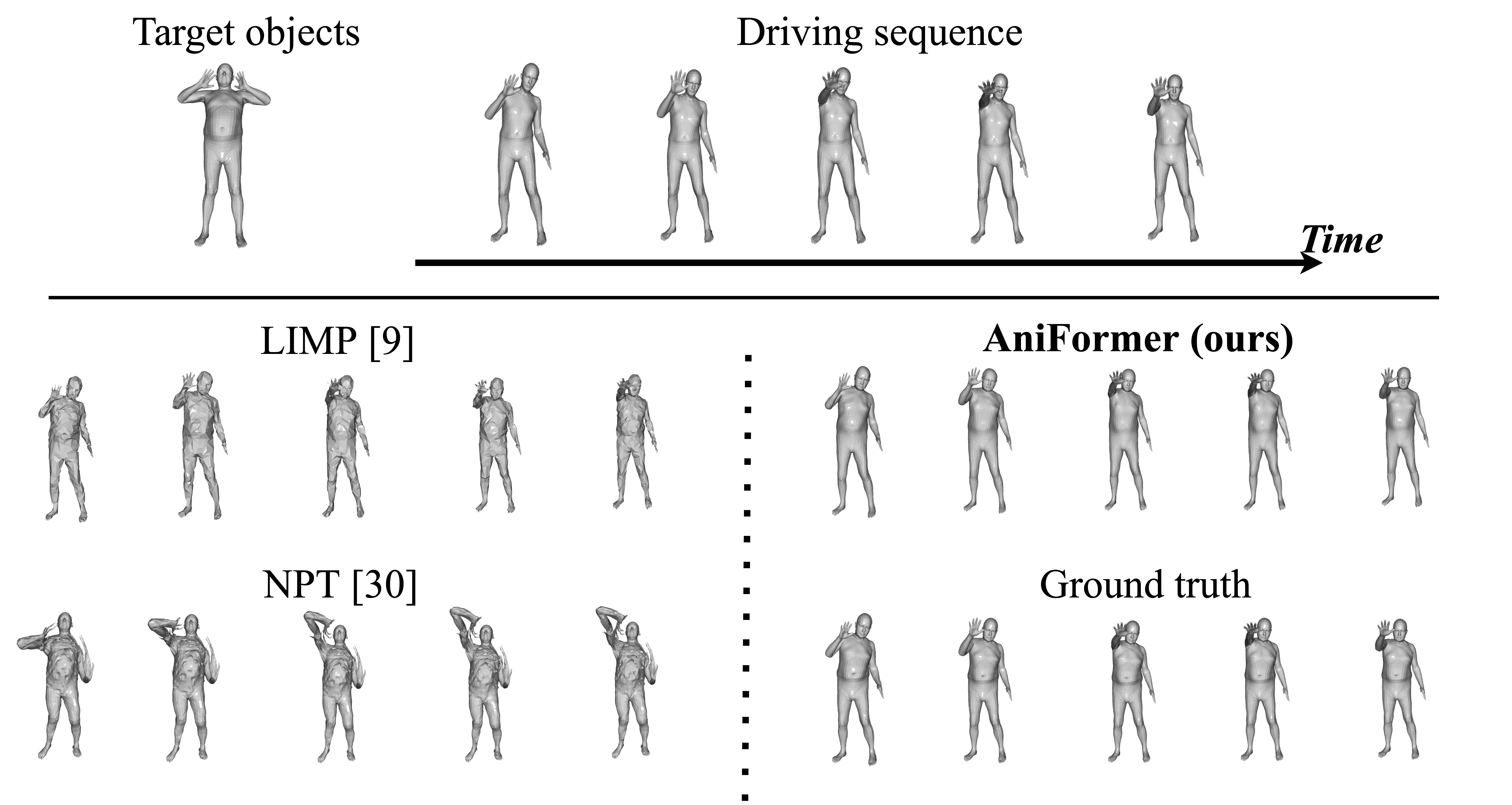}
    \caption{Qualitatively comparison with the state-of-the-art methods on unseen motions from the DFAUST dataset. Given the long driving sequence, compared methods process the deformation frame by frame while AniFormer takes several frames as inputs.}
    \vspace{-0.4cm}
    \label{fig:comparison}
\end{figure}
In this section, we perform an extensive evaluation of proposed AniFormer. In addition to comparing with previous works for evaluating the animating effects, we also analyze the effect of each component in our model.

\noindent\textbf{Datasets.} For training the model, we use the motion sequences in the DFAUST \cite{DFAUST} dataset. DFAUST is a large human motion sequence dataset that captures the 4D motion of 10 human subjects with 129 different body motions, such as ``punching'', ``chicken wings pose'', etc. Each motion consists of hundreds of frames. To create training data, we evenly sample 30 frames from each motion as a complete sequence. Then, one motion (driving sequence) and one appearance (target mesh) will be randomly combined as a pair for training. The ground truth is obtained by using SMPL model \cite{SMPL} to synthesize the target animated sequence with the shape and pose parameters provided by the dataset. The mesh vertices are shuffled randomly to ensure the network is vertex-order invariant.

In testing, we use our model to animate the unseen target meshes with motions from given driving sequences. To do so, we create another 8 new target meshes with the SMPL model by sampled from pose and shape parameter spaces. Then we use the driving sequences to animate these 8 new target meshes. The driving sequences are split into two settings, i.e., the seen driving sequences (80 motions) and the unseen driving sequences (20 motions). Unseen driving sequences are not available in the training procedure. Furthermore, to evaluate the generalization ability of our AniFormer, we employ the model to drive the target meshes from other datasets, e.g., FAUST \cite{FAUST} and MG-dataset \cite{MG-cloth}, for animation.

Lastly, we verify AniFormer on animal meshes from the SMAL dataset \cite{SMAL} via domain-specific training. SMAL is a 3D articulated model that can represent various animals including related animal meshes~\cite{SMAL}. More details of data processing in Supplementary Materials.

\noindent\textbf{Evaluation Metrics.} We introduce the evaluation metrics based on the point-wise mesh Euclidean distance (PMD) which is commonly used as a metric in the 3D pose transfer task. PMD compares the output mesh with the ground truth in a point-wise aligned way. Specifically, we first uniformly sample points from our results and ground truths, then we compute the average PMD over all the frames:
\begin{equation}
\label{PMD}
PMD = \frac{1}{T}\frac{1}{V} \sum_{\mathbf{t}}\sum_{\mathbf{v}}\left \| M^{t}_{\mathbf{v}}-G^{t}_{\mathbf{v}} \right \|_{2}^{2},
\end{equation}
where $M_{\mathbf{v}}$ and $G_{\mathbf{v}}$ are the point pairs from the generated mesh $M$ and ground truth mesh $G$. 

\noindent\textbf{Baselines.} We compare the proposed approach to the most recent single mesh deformation approaches. Specifically, we compare with two state-of-the-art methods, i.e.,
NPT \cite{NPT} and LIMP \cite{LIMP}. Since both methods could only process a single mesh, we use them to generate long sequences frame by frame. We also compare to vanilla Transformer \cite{vit} and its variants. Since so far there is no existing Transformer-based framework for the 3D sequential mesh generation task, we implement them by modifying our AniFormer for a fair comparison. For instance, the regression head version is obtained by replacing the regression constraints of AniFormer back to a classic regression head. Note that, for the compared methods NPT and LIMP, we use the same protocols provided to train the models. For the models in the Ablation study, we trained the compared models within same amount of time to make it fair (also making sure they are converged).

\noindent\textbf{Training and Runtime.} Our network receives a sequence of driving meshes of size $V_1 \times 3 \times T$, and a target mesh of size $V_2 \times 3$ where $V$ and $T$ stand for the vertex number and frame number. Note that $V_1$ might not equal to $V_2$ because our AniFormer can handle meshes of different sizes. The network is implemented in PyTorch \cite{paszke2019pytorch} and optimized using the Adam. The batch size is 2. The total number of training epoch is 200. The learning rate is initialized as 5e-5 and reduced at milestones of 80, 120, and 160 with gamma as 0.1. Due to the memory consumption, the input length of the driving sequences is fixed as 3 frames for both training and testing. The total training time is around 50 hours on an Nvidia GPU V100. During inference, we conduct the animation of the target mesh by using a sliding window with the size of $T$ (3 frames) with the driving sequence as the driving motion. Inference runs at $\sim$170ms per frame with generated meshes of 6,890 vertices.

\noindent\textbf{Ablation Study.} We start the experiments from verifying how much improvement was brought by discarding the regression head. The quantitative and qualitative evaluations are shown in Table~\ref{tab:ablation} and Fig.~\ref{fig:transformer ablation}. We start from a vanilla version with a regression head, then remove the regression head, and lastly add regression constraints. As can be seen, the regression head performs even worse than without head version. One intuitive explanation for this is that, the flattening operator in the MLP blocks will damage the mesh vertex order and the single output lack of temporal consistency with neighboring frames. Particularly, bringing the two regression constraints is very helpful in learning, which means enforcing constraints on multiple outputs could be a better solution for 3D sequential Transformers.

\noindent\textbf{State-of-the-Art Comparisons.} We provide experimental results by comparing AniFormer with state-of-the-art methods both quantitatively in Table~\ref{tab:sota} and qualitatively in Fig.~\ref{fig:comparison}. For both settings: ``Seen motions'' and ``Unseen motions'' from the DFAUST dataset, our AniFormer significantly outperforms compared state-of-the-art methods with frame-level average PMD ($\times 10^{-4}$) of: 0.124 and 3.95 (AniFormer) vs. 1.37 and 11.31 (NPT). Note that LIMP downsamples the meshes from 6,890 to 2,100 vertices to avoid heavy computation while AniFormer processes the original ones. Thus, to make the comparison, we downsample the ground truth meshes to 2,100 vertices when evaluating LIMP to compute the PMD. Fig.~\ref{fig:comparison} shows the qualitative performances of the compared methods on unseen targets on the DFAUST dataset. As shown, given unseen target meshes, both NPT and LIMP suffer from degeneracy, while AniFormer still can provide promising results close to the ground truth.

\begin{figure*}[!t]\small
    \centering
    \includegraphics[width=0.75\textwidth]{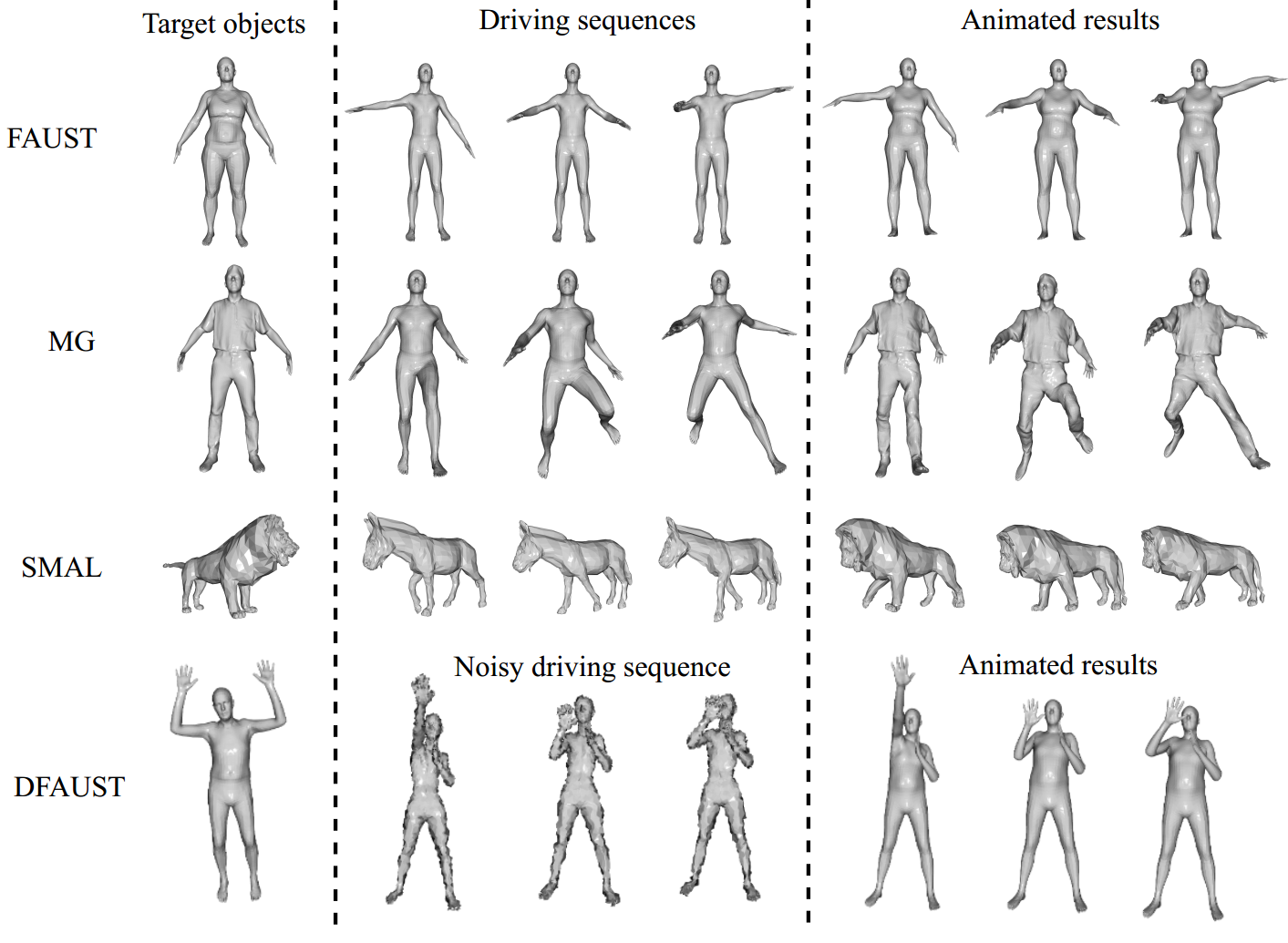}
    \caption{Top three rows: generalization ability of AniFormer on non-SMPL meshes. Last row: Robustness of AniFormer to noisy inputs.}
    \label{fig:ablation}
    \vspace{-0.4cm}
\end{figure*}

\begin{table}[!h]
    \caption{Ablation study on regression head of the Transformer in our AniFormer.}
\centering
\resizebox{0.55\textwidth}{!}{
\begin{tabular}{cc}
\toprule
Method & PMD $\downarrow (\times 10^{-4})$ \\ \midrule
Regression Head (Vanilla) & 15.31 \\
Without Head & 3.84 \\
Motion Constraint & 1.37 \\
Appearance Constraint & 0.31 \\
Full Constraints (AniFormer) & \textbf{0.12} \\ \bottomrule
\end{tabular}
}
\vspace{-0.4cm}
\label{tab:ablation}
\end{table}

\begin{table}[]\small
\centering
  \caption{Comparison with the state-of-the-art methods on the DFAUST datasets. We compute frame-level average point-wise mesh Euclidean distance (PMD) as the metric. Our AniFormer outperforms other compared methods for more than one order of magnitude.}
\resizebox{0.55\textwidth}{!}{
\begin{tabular}{@{}ccc@{}}
\toprule
\multirow{2}{*}{Method} & \multicolumn{2}{c}{PMD$\downarrow (\times 10^{-4})$} \\ \cmidrule(l){2-3} 
 & Seen Motions & Unseen Motions \\ \midrule
LIMP \cite{LIMP} & 5.44 & 7.11 \\
NPT-MP \cite{NPT} & 4.51 & 6.69 \\
NPT \cite{NPT} & 4.37 & 5.31 \\
AniFormer (Ours) & \textbf{0.12} &  \textbf{3.95} \\ \bottomrule
\end{tabular}
}
\vspace{-0.4cm}
\label{tab:sota}
\end{table}

\noindent\textbf{Non-SMPL Mesh Animation.} AniFormer can be generalized to the animation of non-SMPL meshes. As shown in Fig.~\ref{fig:ablation} rows 1-2, unseen meshes from FAUST \cite{FAUST} and MG \cite{MG-cloth} datasets can be directly animated by AniFormer without any further finetune. Note that those meshes are not in line with the SMPL model and more challenging (more fine-grained geometry details). Furthermore, we show the results of our model on animating a lion with the motion of a horse on row 3 of Fig.~\ref{fig:ablation}, by training on the SMAL dataset, showing the capacity of our AniFormer in different domains.

\noindent\textbf{Robustness to Noise.} Furthermore, we evaluate the robustness of AniFormer by adding uniform noise to the vertices of driving meshes. In the last row of Fig.~\ref{fig:ablation}, we can see that AniFormer is insensitive to the noise and can robustly extract the information from driving sequences even with the challenging poses.

\noindent
\textbf{Limitations.} Lastly, we discuss the limitations of AniFormer. Due to the need to preserve the geometric details of the target models as well as model the temporal information, the computational complexity becomes a main issue of AniFormer. Besides, since AniFormer directly uses the motion from the driving sequence to animate the target objects and the driving data from DFAUST is relatively simple, thus the non-linear local deformation of the target objects (plausible subtle muscle moments) has not been explored yet. However, the ability of models to generalize the plausible motions with non-linear local deformation can efficiently improve the animation visual effects which could be future research direction.

\section{Conclusion}
We propose the AniFormer framework, i.e., a novel Transformer network customized for raw-data driven 3D animation with carefully-designed regression constraints. The AniFormer for the first time reveals the potential of animating 3D target meshes via raw driving mesh sequences and qualitative results on various categories further show that our model provides powerful animation ability by maintaining motions depicted in the driving sequences and preserving appearance from target meshes. Future directions could be unsupervisedly learning the motion, improving the AniFormer into a more compact architecture and automatically generalizing the non-linear motion for the target objects.

\noindent \textbf{Acknowledgments.}
This work was supported by the Academy of Finland for project MiGA (grant316765), ICT 2023 project (grant 328115), EU H2020 AI4Media (No.951911) projects and Infotech Oulu, as well as the CSC-IT Center for Science, Finland, for computational resources.


This supplementary material includes technical details and additional results that were not included in the main submission due to the lack of space. All results were obtained with exactly the same methodology as the one described in the main manuscript. We first provide more implementation details of our AniFormer networks (Sec.~\ref{sec:1}). 
Next, we provide more details of processing datasets (Sec.~\ref{sec:2}). 
Finally, detailed training settings are introduced (Sec.~\ref{sec:3}). 

\section{AniFormer Network Architecture}
\label{sec:1}
Our AniFormer framework consists of two main parts: a 3D mesh feature extractor, and an AniFormer Encoder for 3D animation. We first introduce network structures of each component, then give the architectural parameters of the full model.

\noindent \textbf{3D Mesh Feature Extractor}. The architecture of the feature extractor is presented in Table~\ref{Tab:encoder model}. The feature extractors are used to extract a latent embedding of motions from the given mesh sequences for further mesh generation with the following encoders. Note that in order to fit our model on non-SMPL mesh models (the vertex number of which is not equal to 6,890, such as MG-cloth of more than 27,000 vertices), we stack a max pooling layer to the end of the feature extractors. It can flexibly process meshes with different sizes into a certain one. This max pooling version is trained on the DFAUST dataset \cite{DFAUST} (with 6,890 vertices as inputs), then is evaluated on the MG-cloth dataset (with 27,554 as inputs). For the SMAL dataset, we train a model based on it and directly fix the vertex number to 3,889 both both training and testing sets.

\noindent\textbf{AniFormer Encoder}. The network architecture of a AniFormer encoder is presented in Table~\ref{Tab:CGP block}. As mentioned, the MLP blocks in the Vanilla Transformer will damage the mesh vertex order, thus we customize the MLP blocks into an instance normalization (InsNorm) block inspired by \cite{NPT} presented in Table \ref{Tab:SPAdaIN block}. The The AniFormer encoder is used to generate the animated target mesh with given motions.

Finally, we present the full model architecture in Table~\ref{Tab:fullmodel}. The embedded motion features from the driving sequences from the feature extractors will be combined with target meshes and fed into AniFormer encoders and sequential meshes will be generated.

\section{Dataset Settings}
\label{sec:2}
\noindent \textbf{Training Sets.} We use DFAUST dataset \cite{DFAUST} to generate the training dataset. It has 129 motions from ten subjects and each motion last for hundred of frames. Because the motion is captured with high frame rate speed, thus the motion between frames are very subtle. To ensure the the sufficient dynamics between frames, we evenly sample 30 frames from each motion as a complete sequence. Then, one motion (driving sequence) and one appearance (target mesh) will be randomly combined as a pair for training. When training, each time we sample 3 continuous frames of a motion (30 frames) as an driving sequence inputs and feed them to the networks with a random paired target mesh. Since it results in more than 4,000,000,000 potential training pairs (target meshes: 80$\times$ 30 $\times$ 16 with driving sequential meshes: 80$\times$ 30 $\times$ 16$\times$ 27) which is way larger than our computational capacity, we randomly select 8,000 training pairs at each epoch during the training. 

\noindent \textbf{Target Meshes.} Since there are only ten subjects in the DFAUST dataset which is not enough to construct the latent space for appearances. We create 16 meshes for training and another 8 meshes for testing with the SMPL model by randomly sampling from pose and shape parameter spaces. The ground truth is obtained by using SMPL model \cite{SMPL} to synthesize the target animated sequence with the shape and pose parameters provided by the dataset. The mesh vertices are shuffled randomly and the generated faces are correspondingly shuffled to construct the meshes.

\begin{table}[tbp]
	\caption{Detailed architectural parameters for the 3D mesh feature extractor. ``N'' stands for batch size and ``V'' stands for vertex number. The first parameter of Conv1D is the kernel size, the second is the stride size. ``T'' stands for the frame number. The same as below.}
	\centering
	\resizebox{0.5\columnwidth}{!}{%
		\begin{tabular}{@{}cccc@{}}
			\toprule
			Index & Inputs & Operation & Output Shape \\ \midrule
			(1) & - & Input mesh & N×T×3×V \\
			(2) & (1) & Conv1D (1 × 1, 1) & N×T×64×V \\
			(3) & (2) & Instance Norm, Relu & N×T×64×V \\
			(4) & (3) & Conv1D (1 × 1, 1) & N×T×128×V \\
			(5) & (4) & Instance Norm, Relu & N×T×128×V \\
			(6) & (5) & Conv1D (1 × 1, 1) & N×T×1024×V \\
			(7) & (6) & Instance Norm, Relu & N×T×1024×V \\
			(8) & (7) & Max pooling (for non-SMPL) & N×T×1024×V' \\ 
			(9) & (8) & Temporal Embedding & N×T×1024×V \\\bottomrule
		\end{tabular}
	}
	
	\label{Tab:encoder model}
\end{table}

\begin{table}[tbp]
	\caption{Detailed architectural parameters for AniFormer encoder.}
	\centering
	\resizebox{0.5\columnwidth}{!}{%
		\begin{tabular}{cccc}
			\toprule
			Index & Inputs & Operation & Output Shape \\ \midrule
			(1) & - & Driving Motion Embedding & N×C×V \\
			(2) & (1) & Conv1D (1 × 1, 1) & N×T×C×V \\
			(3) & (1) & Conv1D (1 × 1, 1) & N×T×C×V \\
			(4) & (3) & Reshape & N×T×V×C \\
			(5) & (3)(4) & Batch Matrix Product & N×T×V×V \\
			(6) & (5) & Softmax & N×T×V×V \\
			(7) & (6) & Reshape & N×T×V×V \\
			(8) & (1) & Conv1D (1 × 1, 1) & N×T×C×V \\
			(9) & (2)(8) & Batch Matrix Product & N×T×C×V \\
			(10) & (9) & Parameter gamma & N×T×C×V \\
			(11) & (10)(2) & Add & N×T×C×V \\
			(12) & - & Target Mesh & N×3×V \\
			(13) & (11)(12) & InsNorm block & N×T×C×V \\
			(14) & (13) & Conv1D(1 × 1, 1), Relu & N×T×C×V \\
			(15) & (14)(12) & InsNorm block & N×T×C×V \\
			(16) & (15) & Conv1D(1 × 1, 1), Relu & N×T×C×V \\
			(17) & (11)(12) & InsNorm block & N×T×C×V \\
			(18) & (17) & Conv1D(1 × 1, 1), Relu & N×T×C×V \\
			(19) & (15)(18) & Add & N×T×C×V \\ \bottomrule
		\end{tabular}
	}
	
	\label{Tab:CGP block}
\end{table}

\begin{table}[tbp]
	\caption{Detailed architectural parameters for InsNorm block.}
	\centering
	\resizebox{0.5\columnwidth}{!}{%
		\begin{tabular}{cccc}
			\toprule
			Index & Inputs & Operation & Output Shape \\ \midrule
			(1) & - & Driving Motion Embedding & N×T×C×V \\
			(2) & (1) & Instance Norm & N×T×C×V \\
			(3) & - & Target Mesh & N×3×V \\
			(4) & (3) & Conv1D (1 × 1, 1) & N×T×C×V \\
			(5) & (3) & Conv1D (1 × 1, 1) & N×T×C×V \\
			(6) & (4)(2) & Multiply & N×T×C×V \\
			(7) & (6)(5) & Add & N×T×C×V \\ \bottomrule
	\end{tabular}}
	
	\label{Tab:SPAdaIN block}
\end{table}

\begin{table}[tbp]
	\caption{Detailed architectural parameters for the full model.}
	\label{Tab:fullmodel}
	\centering
	\resizebox{0.5\columnwidth}{!}{%
		\begin{tabular}{@{}cccc@{}}
			\toprule
			Index & Inputs & Operation & Output Shape \\ \midrule
			(1) & - & Target Mesh & N×3×V \\
			(2) & - & Driving Motion Mesh & N×T×3×V \\
			(3) & (2) & Feature Extractor & N×T×1024×V \\
			(4) & (3) & Conv1D (1 × 1, 1) & N×T×1024×V \\
			(5) & (4)(1) & AniFormer encoder 1 & N×T×1024×V \\
			(6) & (5) & Conv1D (1 × 1, 1) & N×T×512×V \\
			(7) & (6)(1) & AniFormer encoder 2 & N×T×512×V \\
			(8) & (7) & Conv1D (1 × 1, 1) & N×T×512×V \\
			(9) & (8)(1) & AniFormer encoder 3 & N×T×512×V \\
			(10) & (9) & Conv1D (1 × 1, 1) & N×T×256×V \\
			(11) & (10)(1) & AniFormer encoder 4 & N×T×256×V \\
			(13) & (12) & Conv1D (1 × 1, 1) & N×T×3×V \\
			(14) & (13) & Tanh & N×T×3×V \\ \bottomrule
		\end{tabular}
	}
	
\end{table}

\noindent \textbf{The Driving Sequences} are split into two settings, i.e., the seen driving sequences (80 motions from the first 6 subjects of DFAUST) and the unseen driving sequences (20 motions from the rest 4 subjects of DFAUST). Those 20 motions are only available for evaluating. 

\noindent \textbf{Testing Sets.} For DFAUST dataset, we use those 20 motions mentioned above for testing. Furthermore, we employ the model trained from DFAUST directly to drive the target meshes from other datasets, e.g., FAUST \cite{FAUST} and MG-dataset \cite{MG-cloth}, for animation. As mentioned in the network architecture section, a max pooling layer should be deployed to process the large vertex number (more than 27,000) from MG-cloth dataset. At last, we extent the AniFormer to animal domain on the SMAL dataset \cite{SMAL}. We collect the dataset by pairing the animal templates provided by SMAL and the motion captured by \cite{Zuffi} which uses the SMAL model to generate the 3D posed animals from 2D images.

\section{Experimental Settings}
\label{sec:3}
Our algorithm is implemented in PyTorch~\cite{paszke2019pytorch}. All the experiments are carried out on a PC with a single NVIDIA Tesla V100, 32GB. We train our networks for 200 epochs with a learning rate of 0.00005 and Adam optimizer. The weight settings in the paper are  $\lambda_{rec}{=}1$, $\lambda_{a}{=}0.0005$, and $\lambda_{m}{=}0.0005$. The weight settings directly follow the previous work \cite{NPT}. The batch size is fixed as 2 for all the settings and the frame number is 3. Training time is around 80-90 hours. And the inference time is $\sim$\textit{170 ms} per frame. Note that, batch size of 2 is only available with 32GB memory GPUs to run the AniFormer. For GPUs with 12 or 24GB memory, the batch size should be adjusted to 1.

\bibliography{egbib}
\end{document}